\documentclass[final]{cvpr}

\usepackage{times}
\usepackage{epsfig}
\usepackage{graphicx}
\usepackage{amsmath,bm}
\usepackage{amssymb}
\usepackage{color,xcolor}
\usepackage{xspace}
\usepackage{subfigure}
\usepackage{dirtytalk}

\usepackage[pagebackref=true,breaklinks=true,colorlinks,bookmarks=false]{hyperref}

\def\cvprPaperID{3565} % *** Enter the CVPR Paper ID here
\def\confYear{CVPR 2021}
\begin{document}

%%%%%%%%% TITLE
\title{VERTEX: {VE}hicle {R}econstruction and {TEX}ture Estimation \\Using Deep Implicit Semantic Template Mapping}

\author{
 Xiaochen Zhao\textsuperscript{1},
 Zerong Zheng\textsuperscript{1},
 Chaonan Ji\textsuperscript{1},
 Zhenyi Liu\textsuperscript{2},
 Yirui Luo\textsuperscript{1},
 Tao Yu\textsuperscript{1},
 \\
 Jinli Suo\textsuperscript{1},
 Qionghai Dai\textsuperscript{1},
 Yebin Liu\textsuperscript{1}
 \\
 \textsuperscript{1}Tsinghua University, Beijing, China
 \quad
 \textsuperscript{2}Jilin University, Jilin, China
}

\maketitle

%%%%%%%%% ABSTRACT
\begin{abstract}
We introduce VERTEX, an effective solution to recover 3D shape and intrinsic texture of vehicles from uncalibrated monocular input in real-world street environments. To fully utilize the template prior of vehicles, we propose a novel geometry and texture joint representation, based on implicit semantic template mapping. Compared to existing representations which infer 3D texture distribution, our method explicitly constrains the texture distribution on the 2D surface of the template as well as avoids limitations of fixed resolution and topology. Moreover, by fusing the global and local features together, our approach is capable to generate consistent and detailed texture in both visible and invisible areas. We also contribute a new synthetic dataset containing 830 elaborate textured car models labeled with sparse key points and rendered using Physically Based Rendering (PBRT) system with measured HDRI skymaps to obtain highly realistic images. Experiments demonstrate the superior performance of our approach on both testing dataset and in-the-wild images. Furthermore, the presented technique enables additional applications such as 3D vehicle texture transfer and material identification.

%We believe that VERTEX provides a promising solution for vehicle digitization based on the simplest setup.

%Practically, in order to implement a meaningful semantic template mapping process, the network is trained with a new training strategy to end-to-end optimize shape and texture simultaneously and weak-supervised loss terms. To this end, 

\end{abstract}

%%%%%%%%% BODY TEXT
\section{Introduction}
\label{sec:intro}

% Zerong's revision: 
Monocular visual scene understanding is  a fundamental technology for many automatic applications, especially in the field of autonomous driving. Using only a single-view driving image, available vehicle parsing studies have covered popular topics starting from 2D vehicle detection~\cite{bochkovskiy2020yolov4, Liu_2016_ssd, Lin_2017_ICCV_Foc, Duan_2019_ICCV_CenterNet, Law_2018_ECCV_CornerNet}, then 6D vehicle pose recovery~\cite{Wu_2019_CVPR_Workshops_6DVNet, Single_Image_3D, Kehl_2017_ICCV_SSD_6D, Sundermeyer_2018_ECCV_Implicit_3D, do2019real_Realtime_monocular, do2018deep6dpose, Cai_2020_CVPR_reconstruct_locally, liu20196d_6D_Object_Pose_Estimation}, and finally vehicle shape reconstruction~\cite{Ku_2019_CVPR_Monocular3DObjectDetection, Song_2019_CVPR_ApolloCar3D, Henderson2020Leveraging, Kanazawa_2018_ECCV_Learning_Category-Specific, SAMP, Manhardt2018ROI, Zakharov2020Autolabeling}. However, much less efforts are devoted to vehicle texture estimation, even though both humans and autonomous cars heavily rely on the appearance of vehicles to perceive surroundings. 
% It is well known that human vision relies heavily on the appearance to perceive surroundings, and we are expecting autonomous cars to have the same perceptual power. Hence, it is critical to further estimate the 3D texture of vehicles in street environments. 
Simultaneously recovering the geometry and texture of vehicles is also important for synthetic driving data generation~\cite{li2019aads}, vehicle tracking~\cite{meng2020parsing}, vehicle parsing~\cite{miao2021robust} and so on. 

% Old version: 
% The term “Monocular Visual Scene Understanding” has been used in computer vision to broadly describe high-level understanding of single image content. With the advances on monocular scene understanding in the field of autonomous driving, the community has gradually evolved to deliver finer-grained results, from initial 2D object detection [ ] limited on the image plane, to 3D object detection [ ], which only contain coarse information on the object poses and sizes, to recent research on vehicle shape reconstruction [ ]. It’s known that human vision relies heavily on the appearance of objects to perceive surroundings. Hence, for autonomous vehicles, it is critical to further understand 3D texture of other cars in street environments. Simultaneously recovering the geometry and texture of the vehicle is important for the current research on 3D coherent synthetic data augmentation, dynamic tracking, vehicle ReID and human-machine interaction in the field of autonomous driving. However,  compared with rich research on the instance-level 3D detection and shape recovery, relatively little work has been done to reconstruct 3D texture of vehicles.

\begin{figure}[t]
\begin{center}
\includegraphics[scale=0.31]{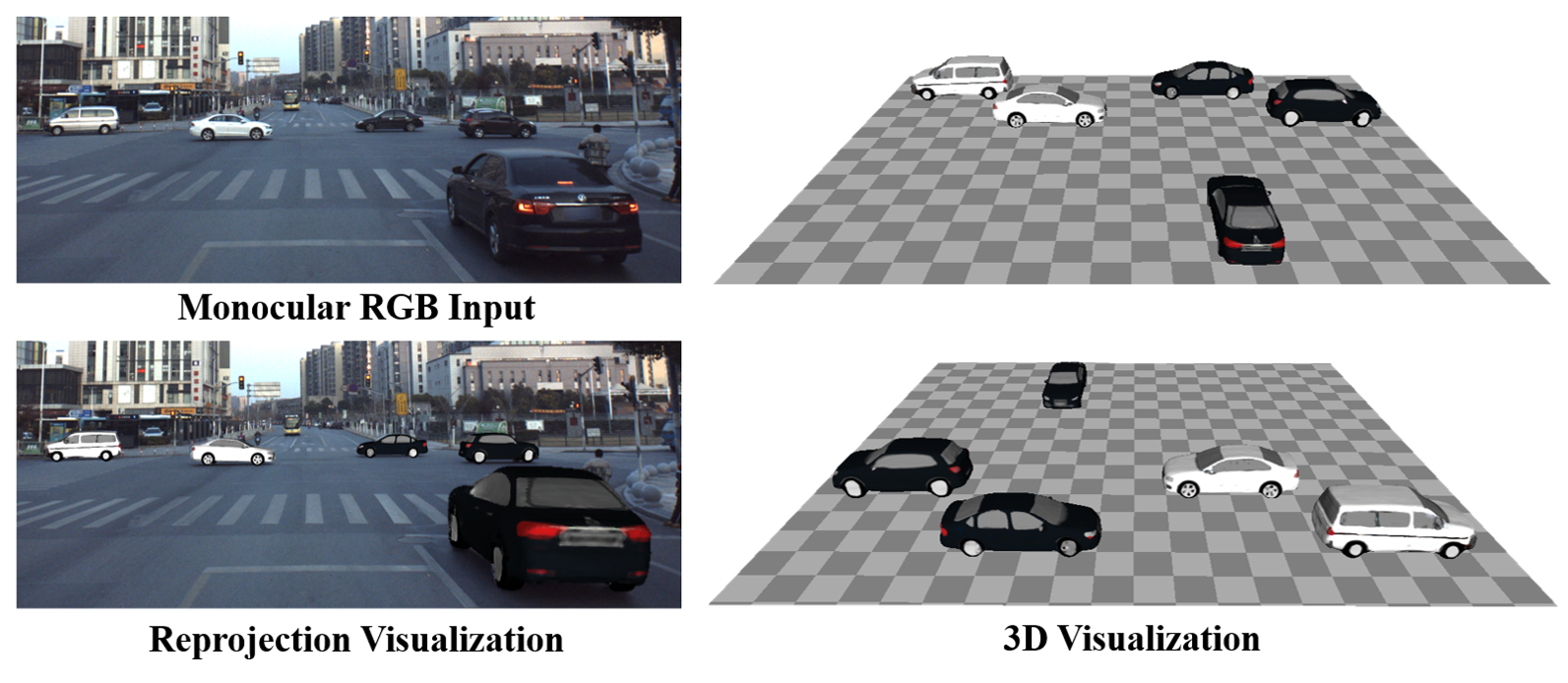}
\end{center}
   \caption{We propose a method to recover realistic 3D textured models of vehicles from a single image (top left) under real street environments. Our approach can reconstruct the shape and texture with fine details. (We manually adjust the scale and layout of models for better visualization.)}
   %(Note that we adjust scale of models for 3D visualization)
\label{fig:uvw_mapping}
\end{figure}

% Zerong's revision:
Challenges for monocular geometry and texture recovery of vehicles mainly arise from the difficulties in inferring the invisible texture conditioned on only  visible pixels while handling various vehicle shapes. Additionally, in real-world street environments, reconstruction methods are also expected to offset the adverse impact of complicated lighting conditions (e.g., strong sunlight and shadows) and diverse materials (e.g., transparent or reflective non-Lambertian surfaces). 
That said, the shape and appearance of vehicles are not completely arbitrary. Our key insight is that those challenges can be addressed with the prior knowledge from vehicle models, especially the part semantics. Therefore, we seek to find a method that is a) aware of the underlying semantics of vehicles, and b) flexible enough to recover various geometric structures and texture patterns.

Recently, deep implicit functions (DIFs), which model 3D shapes using continuous functions in 3D space, have been proven powerful in representing complex geometric structures~\cite{Park2019DeepSDF,Mescheder2019Occupancy}. Texture fields (TF)~\cite{Oechsle2019Texture} and PIFu~\cite{2020PIFu} took a step further by representing mesh texture with implicit functions and estimating point color conditioned on the input image. 
To do so, both TF and PIFu diffuse the surface color into the 3D space. However, it remains physically unclear how to define and interpret the color value off the surface. What's worse, geometry and texture are not fully disentangled in either PIFu or TF, as they rely on the location of surface to diffuse the color into the 3D space, making it difficult to incorporate semantic constraints. 
%Consequently, each texture field can only corresponds to one specific surface. This could be problematic when surface geometry is not available and has to be inferred. If the inferred geometry is slightly different from the ground-truth, the surface texture extracted from the texture field could be erroneous. 

% as one kind of 3D model representation, have attracted much attention in the 3D computer vision community\cite{}. DIFs utilize a neural network to approximate a 3D distribution to represent the information in 3D space. Lots of papers uses DIFs to regress signed distance or occupancy probability to represent 3D shape, showing expressive and flexible capacity for . 
% Although TFs can predict consistent texture around the whole vehicle, the performance of TFs is neither robust nor accurate because the method only embeds global information and infers texture in the whole 3D domain.

In this paper, we explore a novel method, VERTEX, for {VE}hicle {R}econstruction and {TEX}ture estimation from a single image in real-world street environments. 
% To this end, we need a novel representation that can represent both vehicle geometry and texture effectively. 
At its core is a novel implicit geo-tex representation that extends DIFs and jointly represents vehicle surface geometry and texture using implicit semantic template mapping. The key idea is to map each vehicle instance to a canonical template field~\cite{Zheng2021DIT,Deng2021DIF} in a semantic-preserving manner. 
In our geo-tex representation, texture inference is constrained on the 2-manifold of the canonical template; in this way, we can leverage the semantic prior of vehicle template, encourage the model to learn a consistent latent space for all vehicles and bypass the unclear physical meaning of a texture field. 

However, training such a representation for vehicle reconstruction is not straight-forward, because we have no access to the ground-truth mapping from vehicle instances to the canonical template field. 
\cite{Zheng2021DIT,Deng2021DIF} proposed to train the mapping network in an unsupervised manner, and the mapping follows the principle of shortest distance. As a result, the mapping in these methods is not guaranteed to preserve accurate semantic correspondences. 
To resolve this drawback, we propose a joint training method for the geometry reconstruction and texture estimation networks. Our training method is largely different from the training schedule of ``first geometry then texture'' adopted by typical reconstruction works~\cite{2020PIFu, Oechsle2019Texture, Henderson2020Leveraging}. This stems from the insight that the surface texture is closely related to its semantic labels; consider the appearance difference between different parts such as car bodies, windows, tires and lights as examples. The texture information can serve as the additional supervision to force the template mapping to be semantic-preserving .

Trained with our joint training method, our implicit geo-tex representation owns the advantages of both mesh templates and implicit functions: on one hand, it is expressive to represent various shapes, which is the main advantage of DIFs; on the other hand, it disentangles texture representation from geometry, thus supports many downstream tasks including material editing and texture transfer. 
% it is flexible to recover material information for reconstructed vehicle instance by transferring diverse material parameters from parts of a pre-designed template model. Furthermore, our method disentangles the texture and geometry and enables texture transfer between different vehicle shapes, producing semantic meaningful vehicle editing and generation results. 
Although it is initially designed for vehicles, our method can generalize to other objects such as bikes, planes and sofas. 

To simulate real street environments and evaluate our method, we also contribute a synthetic dataset containing 830 elaborately textured car models rendered using Physically Based Rendering (PBRT) system with measured HDRI skymaps to obtain highly realistic images. Each instance is labeled with key points as semantic annotations and can be exploited for evaluation and future research.

% To balance the robustness and accuracy of vehicle texture reconstruction, we combine multi-scale information extracted from the monocular input. Global information is favorable to the reconstruction of overall stable 3D texture, while local information helps to recover fine details. By fusing global and local features, our method could inference stable consistent 3D texture while preserving local details.

% CVPR version
% Our implicit geo-tex representation owns both advantages of the mesh template prior and implicit functions representation for texture inference and is capable of difficult tasks. With the template prior, it is flexible to recover material information for reconstructed vehicle instance by transferring diverse material parameters from parts of a pre-designed template model. Furthermore, our method disentangles the texture and geometry and enables texture transfer between different vehicle shapes, producing semantic meaningful vehicle editing and generation results. 

In summary, our contributions include:
\begin{itemize}
\setlength{\itemsep}{0pt}
\setlength{\parsep}{0pt}
\setlength{\parskip}{0pt}
\vspace{-0.2cm}
    % \item the first method to exploit implicit semantic templates for monocular 3D reconstruction, allowing the disentanglement between volumetrically defined geometry and 2-manifold defined texture;
    \item a novel implicit geo-tex representation with semantic dense correspondences and latent space disentanglement, enabling fine-grained texture estimation, part-level understanding and vehicle editing;
    \item a joint training strategy leveraging the consistency between RGB color and part semantics for semantics-preserving template mapping;
    % \item We experimentally validate that our approach is able to recover realistic vehicle reconstruction from a monocular uncalibrated wild image.
    \item a new vehicle dataset, containing diverse detailed car CAD models, PBRT based rendered images and corresponding real-world HDRI skymaps. 
\end{itemize}

\begin{figure*}
\begin{center}
\includegraphics[scale=0.6]{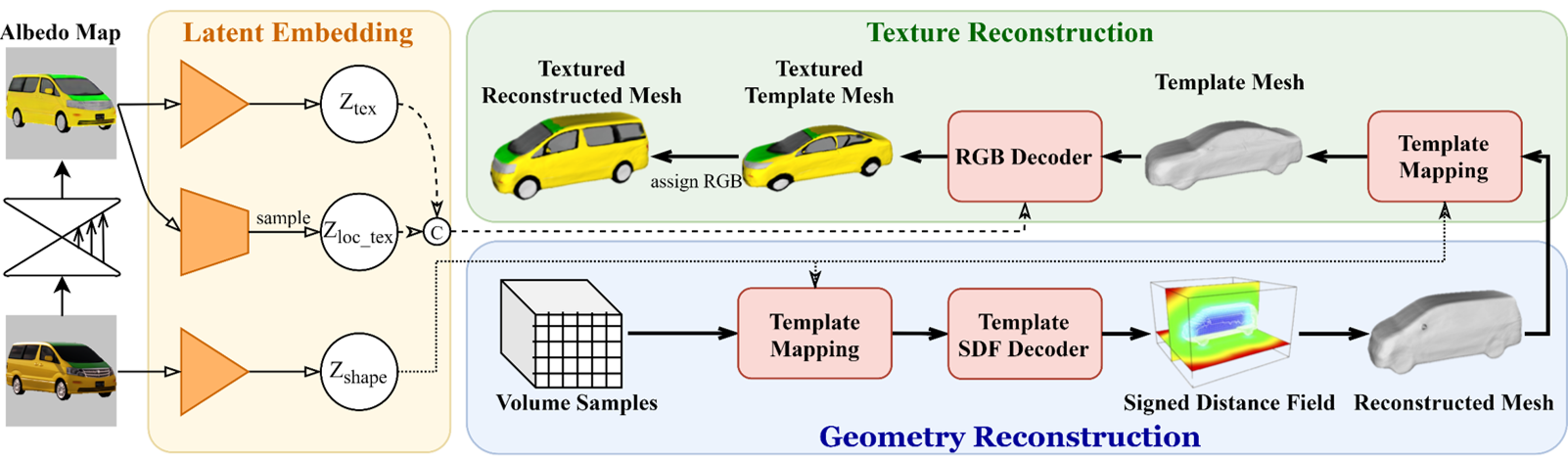}
\end{center}
\caption{The overview of our approach. Given the single RGB image, vehicle digitization is achieved by geometry and texture reconstruction. We first convert the original picture into an albedo map, and then extract multi-scale latent codes in Latent Embedding. Conditioned on these latent codes, our neural networks can infer SDF to reconstruct mesh surface and then regress RGB value for the surface.}
\label{fig:pipeline}
\end{figure*}

%------------------------------------------------------------------------
\section{Related Work}
\label{sec:related}

%-------------------------------------------------------------------------
\subsection{Monocular Vehicle Reconstruction}

In the field of autonomous driving, works for shape recovery and pose estimation~\cite{Kundu_2018_CVPR_3DRCNN,Song_2019_CVPR_ApolloCar3D,Manhardt2018ROI,Zakharov2020Autolabeling,Ku_2019_CVPR_Monocular3DObjectDetection,Grabner_2018_CVPR_3DPoseEstimation,Chabot_2017_CVPR_DeepManta,Mottaghi_2015_CVPR_ACoarse2Fine} can be naively extended to texture reconstruction with projective texturing. However, direct unprojection can only obtain the texture of visible parts and is incapable of recovering consistent 3D texture. 
% For example, ROI-10D[] proposed a monocular deep network that can lift 2D detections in 3D for metrically pose estimation and shape recovery. 

Recently, many works~\cite{beker2020monocularDifferentiableRendering,Henderson2020Leveraging,Kanazawa_2018_ECCV_Learning_Category-Specific, ucmrGoel20} concentrate on vehicle 3D texture recovery under real environments. Due to the lack of ground truth 3D data of real scenes, they mainly focus on the reconstruction from collections of 2D images utilizing unsupervised or self-supervised learning and build on mesh representation. %\cite{beker2020monocularDifferentiableRendering} and \cite{Henderson2020Leveraging} present self-supervised approaches based on differentiable rendering.
%and localization of rigid objects from monocular images.  propose a self-supervised approach that allows learning from collections of 2D images without any 3D information and a new generation process for 3D meshes that guarantees no self-intersections arise.
%\cite{Kanazawa_2018_ECCV_Learning_Category-Specific} is proposed to deform a learned category-specific mean shape with instance-specific predicted offsets. 
Though eliminating the need for 3D annotations and generating meaningful vehicle textured models, these works still suffer from coarse reconstruction results and the limitation of fixed-topology representation. With large-scale synthetic datasets such as ShapeNet~\cite{Chang2015ShapeNet}, many works~\cite{Oechsle2019Texture,sun2018im2avatar,3D-R2N2} train deep neural networks to perform vehicle reconstruction from images. Based on volumentrically representation like 3D voxel~\cite{sun2018im2avatar} and implicit functions~\cite{Oechsle2019Texture}, these works generate plausible textured models in the synthetic dataset, but still struggle with low-quality texture. In contrast, our approach outperforms state-of-the-art methods in terms of visual fidelity and 3D consistency while representing topology-varying objects.

In addition, some works~\cite{VON,Novel_Views_of_Objects, View_synthesis_by_appearance_flow, 2017Transformation} focus on novel view synthesis, i.e., inferring texture in 2D domain. Although they can produce realistic images, they lack compact 3D representation, which is not in line with our goal.

%-------------------------------------------------------------------------
\subsection{Deep Implicit representation}
Traditionally, implicit functions represent shapes by constructing a continuous volumetric field and embed meshes as its iso-surface~\cite{Carr2001Reconstruction,GREG2002Modelling,2004Interpolating}. In recent years, implicit functions have been implemented with neural networks ~\cite{Park2019DeepSDF,Mescheder2019Occupancy,Chen2019Learning,Gropp2020Implicit,2019DISN,2020PIFu,Jiang2020Local, Chabra2020Deep, Hao2020DualSDF} and have shown promising results. For example, DeepSDF~\cite{Park2019DeepSDF} proposed to learn an implicit function where the network output represents the signed distance of the point to its nearest surface. Other approaches define the implicit functions as 3D occupancy probability functions and cast shape representation as a point classification problem~\cite{Mescheder2019Occupancy,Chen2019Learning,2019DISN,2020PIFu}. 

As for texture inference, both TF~\cite{Oechsle2019Texture} and PIFu~\cite{2020PIFu} define texture implicitly as a function of 3D positions. The former uses global latent codes separately extracted from input the image and geometry whereas the latter leverages local pixel-aligned features. Compared with the above approaches~\cite{2020PIFu,Oechsle2019Texture} which predict texture distribution in the whole 3D space, our method explicitly constrains the texture distribution on the 2D manifold of the template surface with implicit semantic template mapping.% Furthermore, we fuse both global and local feature embedding to enhance details as well as keep consistency. 
%Experiments in Sec.~\ref{sec:experiments} show convincing comparison results and demonstrate that our representation is more suitable for vehicle digitization.

\subsection{Warping based Geometry Template Learning}
As one of the most recent research hotpots, warping based geometry can be divided into mesh-based methods~\cite{wang2018pixel2mesh, wen2019pixel2mesh++, gupta2020NMF, jiang2020shapeflow} and implicit methods~\cite{jiang2020shapeflow, gupta2020NMF}, according to different representations.~\cite{wang2018pixel2mesh, wen2019pixel2mesh++} generate various objects by deforming Ellipsoid meshes using graph CNN, while ~\cite{jiang2020shapeflow, gupta2020NMF} use neural ODEs for learned shape deformations instead. Recent works of \cite{Zheng2021DIT, Deng2021DIF} argued for implicitly learning templates for a collection of shapes and establishing dense correspondences. However, without taking texture information into account, the learned geometry templates and the mapping ignore semantic alignment as shown in Fig.~\ref{fig:cmp_trans}. In contrast, our representation presents the semantic template and firstly exploits implicit template prior for monocular 3D reconstruction, achieving good performance.
%-------------------------------------------------------------------------
\section{Implicit Geo-Tex Representation}
\label{sec:overview}
Our method for vehicle reconstruction and texture estimation is built upon a novel geo-tex joint representation, which is presented in this section. 

\subsection{Basic Formulation}
State-of-the-art deep implicit representations for 3D objects, such as PIFu and TF, represent texture and geometry using \emph{separate} implicit fields. However, geometry and texture are never fully disentangled in either PIFu or TF, as they rely on the location of surface to diffuse the color into 3D space. It leads to the fact that each texture field can only corresponds to one specific surface. This could be problematic when surface geometry is not available and has to be inferred. If the inferred geometry is slightly different from the ground-truth, the surface texture extracted from the texture field could be erroneous.

%In order to represent texture in an implicit manner, both TF and PIFu diffuse the surface texture into 3D space. However, it remains physically unclear how to define and interpret the color value off the surface. What's worse, geometry and texture are never fully disentangled in either PIFu or TF, as both PIFu and TF rely on the location of surface to diffuse the color into 3D space. Consequently, each texture field can only corresponds to one specific surface. This could be problematic when surface geometry is not available and has to be inferred. If the inferred geometry is slightly different from the ground-truth, the surface texture extracted from the texture field could be erroneous. 

% In order to decouple the texture from the geometry, both TF\cite{Oechsle2019Texture} and PIFu\cite{2020PIFu} try to diffuse the surface texture into 3D space. 
% However, we argue that this kind of texture representation is physically ambiguous because the meaning of the color values off the surface is unclear. 
We believe that an ideal geo-tex representation should disentangle texture representation from geometry as uv mapping does and should be accord with the physical fact that texture only attaches to the 2D surface of the object. In particular, observing that vehicles are a class of objects with a strong template prior, we extend DIT~\cite{Zheng2021DIT} and propose a \emph{joint} geo-tex representation using deep implicit semantic templates. The key idea is to manipulate the implicit field of the vehicle template to represent vehicle geometry while embedding texture on the 2-manifold of the template surface. Mathematically, we denote the vehicle template surface with $\mathcal{S}_T$ as the level set of a signed distance function $F: \mathbb{R}^3\mapsto\mathbb{R}$, i.e. $ F(\bm{q})=0$, where $\bm{q}\in\mathbb{R}^3$ denotes a 3D point. Then our representation can be formulated as:
% For other implicit functions based geometry-texture representation, they follow the principle “texture after geometry” and they embed surface texture as a continuous function in 3D space. Different from this formulation, we hold the view that, . For vehicles, a class of objects with strong template prior, our method adopt a shape-texture joint representation by means of implicit semantic template, which can be formulated as below:
\begin{equation}
\left\{
    \begin{array}{l}
        \bm{p}_{tp}=W(\bm{p}, \bm{z}_{shape})\\
        s=F(\bm{p}_{tp})\\
        \bm{p}_{tp}^{(S)}=W(\bm{p}^{(S)}, \bm{z}_{shape})\\
        c=T(\bm{p}_{tp}^{(S)}, \bm{z}_{tex})
    \end{array}
\right. 
\end{equation}
where $W:\mathbb{R}^3\times\mathcal{X}_{shape}\mapsto\mathbb{R}^3$ is a spatial warping function mapping the 3D point $\bm{p}\in\mathbb{R}^3$ to the corresponding location $\bm{p}_{tp}$ in the canonical template space conditioned on the shape latent code $\bm{z}_{shape}$, and $F$ queries the signed distance value $s$ at $p_{tp}$. $\bm{p}^{(S)}\in\mathcal{S}\subset\mathbb{R}^3$ is a 3D point on the vehicle surface $\mathcal{S}$, which is also mapped onto the template surface $\mathcal{S}_T$ using the warping function $W$, 
and $T:\mathcal{S}_T\times\mathcal{X}_{tex}\mapsto\mathbb{R}^3$ regresses the color value $c$ of the template surface point  $\bm{p}_{tp}^{(S)}$ conditioned on the texture latent code $\bm{z}_{tex}$. Intuitively, we map the vehicle surface to the template using warping function $W$ and embed the surface texture of different vehicles onto one unified template. Therefore, in our representation, texture is only defined on the template surface (a 2D manifold), avoiding unclear physical meaning of a three-dimensional texture field.

% \begin{figure}[t]
% \begin{center}
% \includegraphics[scale=0.39]{template_mapping.png}
% \end{center}
%   \caption{Visualization of implicit semantic template. With template as an intermediary, our representation is able builds up dense correspondences (color-coded shown in left) and preserve semantic mapping (right).}
% \label{fig:uvw_mapping}
% \end{figure}

\subsection{Formulation for Image-based Reconstruction}
For a specific instance, the shape information is defined by $\bm{z}_{shape}$, while the texture information is encoded as $\bm{z}_{tex}$, both of which can be extracted from the input image using CNN-based encoders. 
%Due to the disentanglement of texture prediction on the 2-manifold, we can reconstruct realistic textured vehicle models compared with state-of-the-art methods. 
To further preserve fine details presented in the monocular observation, we fuse local texture information represented as $\bm{z}_{loc\_tex}(\bm{p})$ at the pixel level. Not only the texture in visible region can benefit from local features, invisible regions can also be enhanced with the structure prior of the template. 
% CVPR version
%For a specific instance, the shape information is defined by $\bm{z}_{shape}$, determining the mapping from the instance space to the template space, while the texture information is encoded as $\bm{z}_{tex}$, determining texture pattern on the template surface. Both $\bm{z}_{shape}$ and $\bm{z}_{tex}$ can be extracted from the input image using CNN-based image encoders. Unfortunately, although the global latent codes contribute to consistent 3D texture estimation in unseen regions, recovered texture tends to be over-smooth and lacking fine details. To overcome this challenge, we fuse $\bm{z}_{tex}$ with \emph{local} feature representation $\bm{z}_{loc\_tex}(\bm{p})$ at the pixel level to preserve the local detail present in the image.  Not only the texture in visible region can benefit form local features, invisible regions can also be enhanced using the symmetry of vehicle. 
% It’s worth mentioning that our representation is competent for uncalibrated in-the-wild images while preserving fine details using local features, by setting a virtual camera to get rid of the request on ground truth camera parameters. 
Formally, our formulation can be rewritten as:
\begin{equation}
\left\{
    \begin{array}{l}
        \bm{p}_{tp}=\mathcal{W}(\bm{p}, \bm{z}_{shape})\\
        s=F(\bm{p}_{tp})\\
        \bm{p}_{tp}^{(S)}=W(\bm{p}^{(S)}, \bm{z}_{shape})\\
        c=T(\bm{p}_{tp}^{(S)},\bm{z}_{tex},\bm{z}_{loc\_tex}(\bm{p}))
    \end{array}
\right. 
\end{equation}
where $T:\mathcal{S}_T\times\mathcal{X}_{tex}\times\mathcal{X}_{loc\_tex}\mapsto\mathbb{R}^3$ is conditioned on the latent codes $\bm{z}_{tex} and \bm{z}_{loc\_tex}$ .

Compared with the previous works, the main advantage of our joint representation is that it explicitly constrains the texture distribution on the 2D surface of the template model, which effectively reduces the complexity of regressing texture. Besides, with the template being an intermediary, shape latent codes and texture latent codes are well decoupled. As a result, it is easy to combine different pairs of latent codes to transfer texture across shapes, as demonstrated in Fig.~\ref{fig:tex_trans}. Moreover, template can be custom-designed to assign extra semantics, such as material information. Observing that vehicles always share similar material in corresponding parts (e.g. glass in car window, metal in car body), our representation can become a promising solution to monocular vehicle material recovery. 
%the introduction of implicit semantic template also has many additional benefits. 

In summary, aiming at vehicle texture recovery, our representation is more expressive with less complexity. However, implementing and training our representation for textured vehicle reconstruction is not straight-forward. We will introduce how we achieve this goal in Section~\ref{sec:method}.

\begin{figure*}[t]
\begin{center}
\includegraphics[scale=0.4]{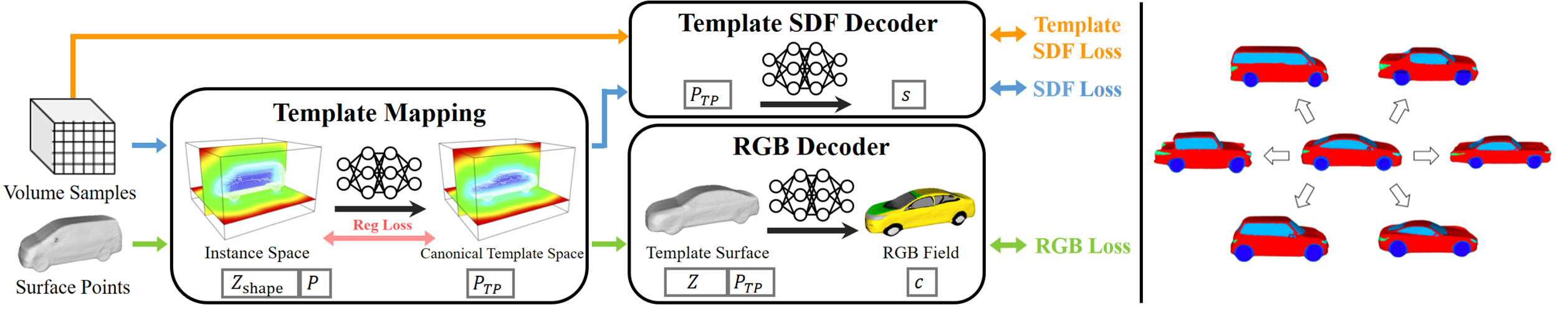}
\end{center}
\caption{To implement implicit semantic template mapping (right), we minimize both data terms of geometry (blue arrows) and texture (green arrows) reconstruction simultaneously. Besides, the regularization terms (orange and pink arrows) for specific network modules are applied to assist training. Each neural network module plays the role of domain mapping conditioned on corresponding latent codes as mentioned in overview. Note that $Z$ in RGB Decoder is the concatenation of the global and local texture latent codes. }
\label{fig:train_pipe}
\end{figure*}
%-------------------------------------------------------------------------
\section{Joint Geo-tex Training Method}
\label{sec:method}
In this section, we first describe our network architecture in Sec.~\ref{sec:method:architecture}. In Sec.~\ref{sec:method:loss}, we present how we train our geometry reconstruction network and texture estimation network jointly. The inference scheme is presented in Sec.~\ref{sec:method:inference}. 

\subsection{Network Architecture}
\label{sec:method:architecture}
Fig.~\ref{fig:pipeline} illustrates the overview of our network, consisting of three modules, i.e., Latent Embedding (yellow), Geometry Reconstruction (blue) and Texture Estimation (green). Our network takes as input a single vehicle image and corresponding 2D silhouette, which can be produced by off-the-shelf 2D detectors~\cite{2017Mask}, and generates a textured mesh. %Note that unlike DIT \cite{Zheng2021DIT}, our vehicle template model is manually specified using one car CAD model in Sec.\ref{sec:method:dataset}. 

\textbf{Albedo Recovery}:
We empirically found that directly extracting texture latent codes from the input images leads to unsatisfactory results. Therefore, before feeding the input image to our network, we first infer the intrinsic color in 2D domain by means of image-to-image translation~\cite{ronneberger2015unet}, and the recovered albedo image will be used as the input for texture encoders in Latent Embedding. We find this module effectively contributes to alleviating the noise effects of image illumination on consistent texture recovery. 
% As for the problem of recovering texture in the intrinsic domain, we have tried to extract texture latent codes directly from the input image, and ask RGB Decoder to learn to remove complex lighting effects. However, such solution worsens the accuracy of texture estimation and lead to over-smooth results. Considering the good performance of image-to-image translation networks in the 2D field, we train a UNet-based network to generate albedo image.

\textbf{Latent Embedding}:
The global shape and texture latent codes, $\bm{z}_{shape}$ \& $\bm{z}_{tex}$, are extracted from the input image and recovered albedo map using two separate ResNet-based~\cite{ResNet} encoders respectively. The local texture feature, $\bm{z}_{loc\_tex}(\bm{p})$, is sampled following the practice of PIFu~\cite{2020PIFu}. Different with other texture inference works~\cite{2020PIFu,Oechsle2019Texture} which only utilize either global or local features for texture reconstruction, we fuse \emph{multi-scale} texture features to recover robust and detailed texture. 

%As for $\bm{z}_{pose}$, we first estimate the vehicle pose from the input image using another encoder, and then lift the pose to a high-dimensional latent vector using an MLP. Since most vehicles in street view only rotate around the yaw axis, we only estimate and lift the egocentric yaw-orientation~\cite{Kundu_2018_CVPR_3DRCNN} from the image. 

% Similar to most implicit functions based methods[], we use latent codes extracted from images to condition implicit function network blocks. In the latent embedding, from 4-dim input (rgbm) multi ResNet-based[] encoders are used to separately embed shape, pose, global and local texture information as $\bm{z}_{shape}$, $\bm{z}_{tex}$, $\bm{z}_{loc\_tex}$ and $\bm{z}_{pose}$. $\bm{z}_{loc\_tex}$ is sampled according to the 2D projection of $\bm{p}_i$ on the feature maps. Different with other texture inference works[TF,Pifu] which only utilize global or local features for texture reconstruction, we fuse multi-scale texture features to recover robust and detailed texture. 

% As for $\bm{z}_{pose}$, as 3D rotations of most vehicle in street view are mainly focus on the rotation of the yaw axis, we simply represent the pose information with the egocentric orientation~\cite{}[3DRCNN] of the yaw axis. $\bm{z}_{pose}$ is generated by upgrading the dimension with a simple fully-connected net.

\textbf{Geometry Reconstruction \& Texture Estimation}:
These two modules form the core of VERTEX. They consist of three main components: Template Mapping, Template SDF Decoder and RGB Decoder. 
% Both continuous implicit mapping functions represented by multi-layer perceptions(MLPs). 
Conditioned on $\bm{z}_{shape}$, volume samples are sequentially fed to the Template Mapping and Template SDF Decoder to predict the continuous signed distance field. For texture estimation, surface points on reconstructed mesh are firstly warped to the template surface conditioned on $\bm{z}_{shape}$, and then passed through the RGB Decoder with embedding latent codes $\bm{z}_{tex}$, $\bm{z}_{loc\_tex}(\bm{p})$ and $\bm{z}_{pose}$ to predict texture.
%Inspired by recent work \cite{tancik2020fourier} using Fourier encoding, we apply a simple linear layer to mapping $\bm{p}_{tp}$ to high-dimensional vectors, which can obviously accelerate the convergence during training.

%-------------------------------------------------------------------------
\subsection{Network Training}
\label{sec:method:loss}
Based on our implicit geo-tex representation, we train the geometry and texture reconstruction network jointly. In this way, we are able to leverage the consistency between RGB color and semantic part segmentation to force the template mapping to be semantic-preserving. We visualize the training process in Fig.~\ref{fig:train_pipe} and provide detailed definition of our training losses.

\textbf{Data Loss:}
For geometry reconstruction, we mainly train by minimizing the $\mathcal{\ell}1$-loss between the predicted and the ground-truth point SDF values:
\begin{equation}
    L_{geo}=\frac{1}{N_{sdf}}\sum_{i=1}^{N_{sdf}}\left \| T(W\left ( \bm{p}_i, \bm{z}_{shape} \right ))-s_i  \right \|_1 \label{eq7}
\end{equation}
where $N_{sdf}$ represents the number of input sample points, $\bm{z}_{shape}$ is the shape latent code corresponding to the volume sample point $\bm{p}_i$, and $s_i$ is the corresponding ground truth SDF value on the $\bm{p}_i$. 

To train the texture estimation network, we minimize the $\mathcal{\ell}1$-loss between the regressed and the ground-truth intrinsic RGB value:
\begin{equation}
\begin{aligned}
    L_{tex}=\frac{1}{N_{sf}}\sum_{i=1}^{N_{sf}}\left \| T\left (W\left ( \bm{p}_i^{(S)}, \bm{z}_{shape} \right ), \bm{z}_{tex},\right. \right. \\
    \left. \left. \bm{z}_{loc\_tex}(\bm{p}_i^{(S)})\right )-c_i  \right \|_1 \label{eq9}
\end{aligned}
\end{equation}
where $N_{sf}$ represents the number of input surface points, $c_i$ is the corresponding ground truth color value on the surface point $\bm{p}_i$, and $\bm{z}_{shape}$, $\bm{z}_{tex}$ and  $\bm{z}_{loc\_tex}$ are the latent codes corresponding to the $\bm{p}_i^{(S)}$. 
% $F$ is a simple function represented by MLP to fuse global and local texture latent codes. 

\textbf{Regularization Loss:}
To establish continuous mapping between the instance space and the canonical template space, we introduce an additional regularization term to constrain position offsets of points after warping:
\begin{equation}
    L_{reg}=\frac{1}{N_{sdf}}\sum_{i=1}^{N_{sdf}}\left \| W\left ( \bm{p}_i, \bm{z}_{shape} \right )- \bm{p}_i  \right \|_2
    \label{eq10}
\end{equation}

\textbf{Template SDF Supervision:}
We supervise Template SDF Decoder directly using the sample points of the template car model. The loss is defined as:
\begin{equation}
    L_{tp\_sdf}=\frac{1}{N_{tp\_sdf}}\sum_{i=1}^{N_{tp\_sdf}}\left \| T(\bm{p}_i^{(tp)})-s_i^{(tp)}  \right \|_1 
\label{eq8}
\end{equation}
where $N_{tp\_sdf}$ represents the number of input sample points, $\bm{p}_i^{(tp)}$ represents the volume sample point around template model and $s_i^{(tp)}$ is the corresponding SDF value.

% In practice, we pretrain Template SDF Decoder with $L_{tp\_sdf}$. Keeping $L_{tp\_sdf}$ smaller than $L_{geo}$ helps ensure that the output field of Template Mapping should be as close as possible to template filed.

% \textbf{Sparse Correspondence Supervision}
% Keypoint pairs between instance vehicles and template model contain sparse correspondence information and can serve as a weak semantic supervision for the template mapping network. Therefore, in order to obtain semantic template mapping, key points pairs are used to construct a loss as:
% \begin{equation}
%     L_{kps}=\frac{1}{N_{kps}}\sum_{i=1}^{N_{kps}}\left \| W\left (\bm{k}_i, \bm{z}_{shape} \right ) -\bm{k}_i^*  \right \|_1 \label{eq11}
% \end{equation}
% where $N_{kps}$ represents the number of input key points, $\bm{k}_i$ and $\bm{k}_i^*$ are the corresponding keypoints on vehicle instance and the template, respectively. 

\begin{figure*}[t]
\begin{center}
\includegraphics[scale=0.7]{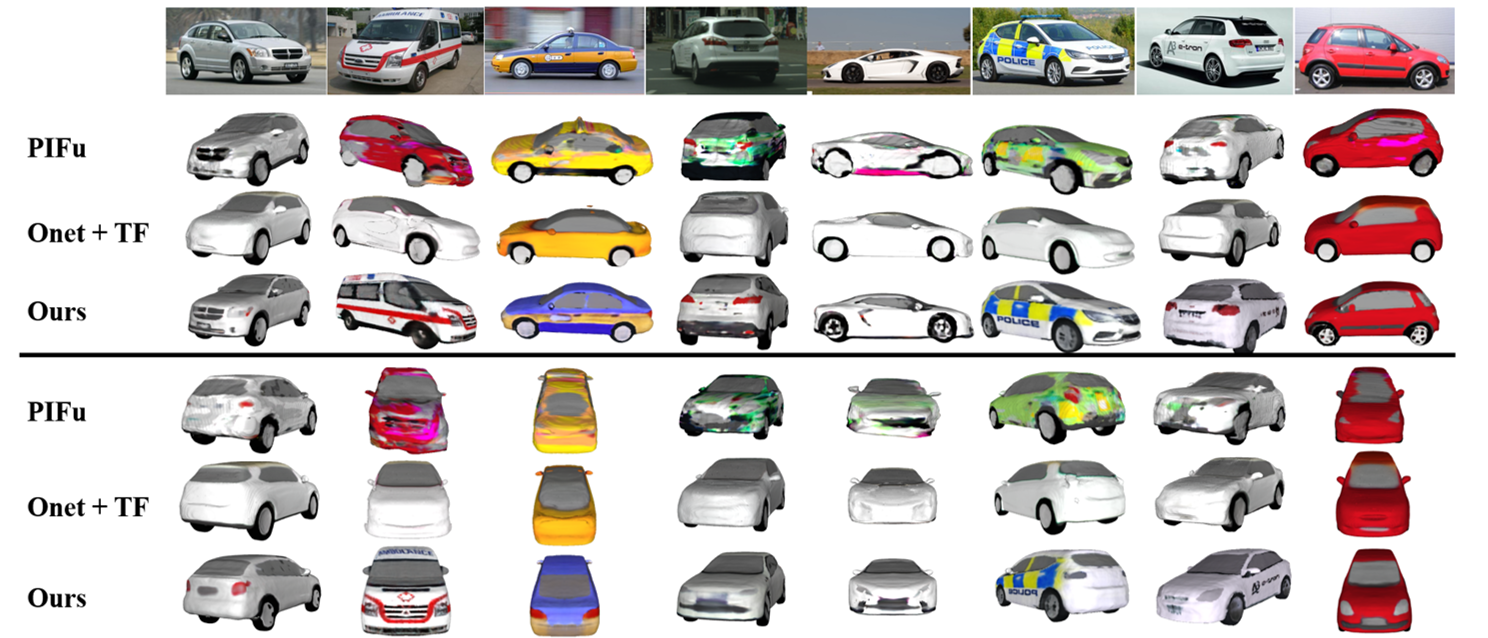}
\end{center}
\caption{Results on in-the-wild images. Monocular input images are shown in the top row. We compare 3D models reconstructed by ours and contrast works (PIFu and Onet+TF) retrained with our dataset. Two render views are provided to demonstrate reconstruction quality. Our results achieve great performance in terms of both robustness and accuracy. }
\label{fig:compare}
\end{figure*}

% \textbf{Pose Loss}
% To supervise the pose encoder, we use 2D sin-cos-encoding vector according to the periodicity of rotation instead of directly using the 1D parameter of yaw rotation:
% \begin{equation}
%     L_{pose}=\left \| o - o^* \right \|_1 \label{eq10}
% \end{equation}
% where $o$ is the sin-cos-encoding of the orientation and $o^{*}$ present the corresponding ground truth. 

% \textbf{Training Scheme}
% Instead of firstly training geometry network and then texture, we simultaneously train both of them using the losses defined above, leveraging the consistency between RGB color and semantic parts to guide the template mapping to be semantic-preserving.

% It is found that jointly training $\bm{z}_{tex}$ and $\bm{z}_{loc\_tex}$ suffers from overfitting. The reason is that it’s much easier for $\bm{z}_{loc\_tex}$ to converge than $\bm{z}_{tex}$ causing the recovered texture to rely on $\bm{z}_{loc\_tex}$ and $\bm{z}_{tex}$ ineffectual, which is harmful to generate robust and consistent texture estimation. Hence, practically we first input $\bm{z}_{tex}$ alone as texture representation by setting $\bm{z}_{loc\_tex}$ as all-zero vectors, and $\bm{z}_{loc\_tex}$ is incorporated into training after basic convergence.

Overall, the total loss function is formulated as the weighted sum of above mentioned terms:
\begin{equation}
    L = L_{tex}+w_{g}L_{geo}+w_{reg}L_{reg}+w_{t}L_{tp\_sdf}
    \label{eq13}
\end{equation}

With embedding latent codes implicitly depending on the parameters of encoders, the whole network is trained end-to-end by minimizing Eq.~\ref{eq13}. See supplementary for implementation details.

%-------------------------------------------------------------------------
\subsection{Inference}
\label{sec:method:inference}

As shown in the pipeline in Fig.~\ref{fig:pipeline}, during inference, we first regress the signed distance field with the branch of geometry reconstruction, and then 3D points on the extracted surface are input to the branch of Texture Estimation to recover surface texture. However, because of the lack of ground truth camera intrinsic and extrinsic parameters, it is difficult for a 3D point to sample the correct local feature from feature map, which poses a significant challenge. We address the problem by setting a virtual camera and further optimizing the 6D pose under the render-and-compare optimization framework. See supplementary for details. 

%-------------------------------------------------------------------------
\section{Experiments}
\label{sec:experiments}
In this section, we first introduce the new vehicle dataset in Sec.~\ref{sec:dataset}. In Sec.~\ref{sec:results}, we illustrate the reconstruction results under real environments and quantitative scores on our dataset compared with two state-of-art baselines. For evaluation in Sec.~\ref{sec:evaluation}, we conducted ablation studies. Finally, we show results on other object categories to prove representation generalization in Sec.~\ref{subsec::Generalization}. More experimental details are presented in the supplementary.
\subsection{Dataset}
\label{sec:dataset}
To generate synthetic dataset, we collect 83 industry-grade 3D CAD models covering common vehicle types, each of which is labeled with 23 semantic key points. These key points pairs contain semantic correspondences and are served to evaluate the accuracy of semantic correspondence in our experiments. We specifically select a commonly seen car as the vehicle template. To enrich the texture diversity of our dataset, we assign ten different texture for each model. To simulate the driving view in real street environment, car models are randomly rotated and placed in different 3D locations, and then rendered in high-resolution (2048$\times$1024) and wide-angle ($fov=50^{\circ}$) image. We generate images with high visual fidelity using Physically Based Rendering (PBRT)~\cite{pbrt} system and measured HDRI skymaps in the Laval HDR Sky Database~\cite{Laval_HDR_sky}. Finally, we get a training set with 6300 instances and a testing set with 2000 instances in total. Please refer to supplementary for more details.

As for the supervision for geometry reconstruction, we use the same data preparation method as Onet~\cite{Mescheder2019Occupancy} to generate watertight meshes and follow the sampling strategy in DeepSDF~\cite{Park2019DeepSDF} to obtain spatial points with their calculated signed distance value.

\subsection{Results and Comparison}
\label{sec:results}
We compare our method with two state-of-the-art methods based on implicit functions. One is PIFu~\cite{2020PIFu} which leverages pixel-aligned features to infer both occupied probabilities and texture distribution. The other one is Onet + Texture Field~\cite{Mescheder2019Occupancy,Oechsle2019Texture}, of which Onet reconstructs shape from the monocular input image and TF infers the color for the surface points conditioned on the image and the geometry. For fair comparison, we retrain botth methods on our dataset by concatenating the RGB image and the instance mask image into a 4-channel RGB-M image as the new input. Specifically, for PIFu, instead of the stacked hourglass network~\cite{2016Stacked} designed for human-related tasks, ResNet34 is set as the encoder backbone and we extract the features before every pooling layers in ResNet to obtain feature embeddings. For Onet and TF, we use the original encoder and decoder networks and adjust the dimensions of the corresponding latent codes to be equal to those in our method.

\textbf{Qualitative Comparison.} To prove that our method adapts to real-world images, we collect several images from Kitti~\cite{Menze2018Kitti}, CityScapes~\cite{Cordts2016CityScapes}, ApolloScape~\cite{wang2019apolloscape}, CCPD\footnote{https://github.com/nicolas-gervais/predicting-car-price-from-scraped-data/tree/master/picture-scraper}, SCD~\cite{KrauseStarkDengFei-Fei_3DRR2013} and Internet. As shown in Fig.~\ref{fig:compare}, our approach generates more robust results when compared with PIFu, while recovering much more texture details than the combination of Onet and TextureField. 

\begin{table}
\begin{center}
\begin{tabular}{lll}
\hline
Method & FID {$\downarrow$} & SSIM {$\uparrow$} \\
\hline
PIFu* & 215.8 & 0.6962 \\
Onet+TF* & 262.73 & 0.7002 \\
\hline
Ours(w/o local feature fusion) &156.8  &0.7057  \\
Ours & \textbf{148.2} & \textbf{0.7208} \\
\hline
Ours(w/o joint training) &193.6   &0.6902  \\
\hline
Ours(MPV as the template) &173.2 &0.6895  \\
Ours(coupe as the template) &159.7  &0.6983  \\
Ours(sphere as the template) &187.4  &0.6833  \\
\hline
\end{tabular}
\end{center}
\caption{Quantitative Evaluation using the FID and SSIM metrics on our dataset. For SSIM, larger is better; for FID, smaller is better. Our method achieves best in both two terms.}
\label{table:idx}
\end{table}

\begin{figure*}[t]
\begin{center}
\includegraphics[scale=0.5]{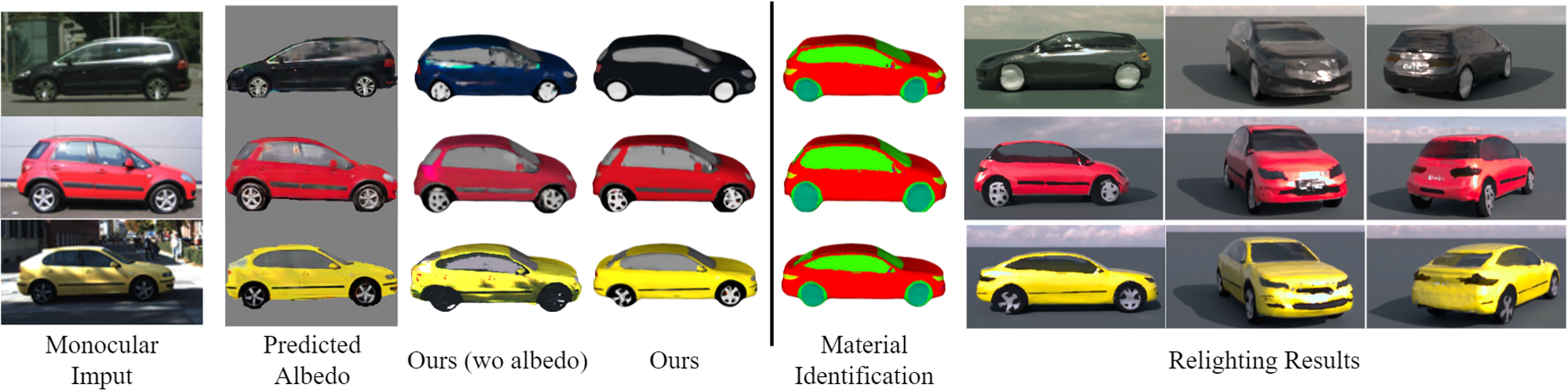}
\end{center}
\caption{Illumination Removal and Material Analysis. Conditioning texture inference with the predicted albedo maps improves the reconstruction robustness. Then, benefiting from our implicit semantic template mapping, we can assign material information from the pre-designed template for the reconstructed model and then render realistic images. Note that different specular reflections shown in right cols are caused by material differences (e.g. metal material body and glass material window), proving our diverse material identification.}
\label{fig:de-illumination}
\end{figure*}

\textbf{Quantitative Comparison.}
To quantitatively evaluate the reconstruction quality of different methods, we use two metrics: Structure similarity image metric (SSIM)~\cite{SSIM} and Frechet inception distance (FID)~\cite{FID}. These two metrics can respectively measure local and global quality of images. The SSIM is a local score that measures the distance between the rendered image and the ground truth on a per-instance basis (larger is better). FID is widely used in the GAN evaluation to evaluate perceptual distributions between a predicted image and ground truth. It is worth noting that both SSIM and FID can not evaluate the quality of generated texture of 3D objects directly. All textured 3D objects must be rendered into 2D images from the same viewpoints of ground truth. To get a more convincing result, for each generated 3D textured model, we render it from 10 different views and evaluate the scores between renderings and corresponding ground truth albedo images. As shown in Tab.~\ref{table:idx}, our method gives significantly better results in FID term and achieves state-of-the art result in SSIM term, proving that our 3D models preserve stable and fine details under multi-view observations. The quantitative results agree with the performance illustrated in qualitative comparison.

We also implement a variant of our method which does not fuse local features for the purpose of fair comparison. As shown in Tab.~\ref{table:idx}, our reconstruction conditioned on global latent codes still outperforms 'Onet+TF', demonstrating that our representation is more expressive in terms of inferring the texture on the vehicle surface.

\begin{figure}
\begin{center}
\includegraphics[scale=0.38]{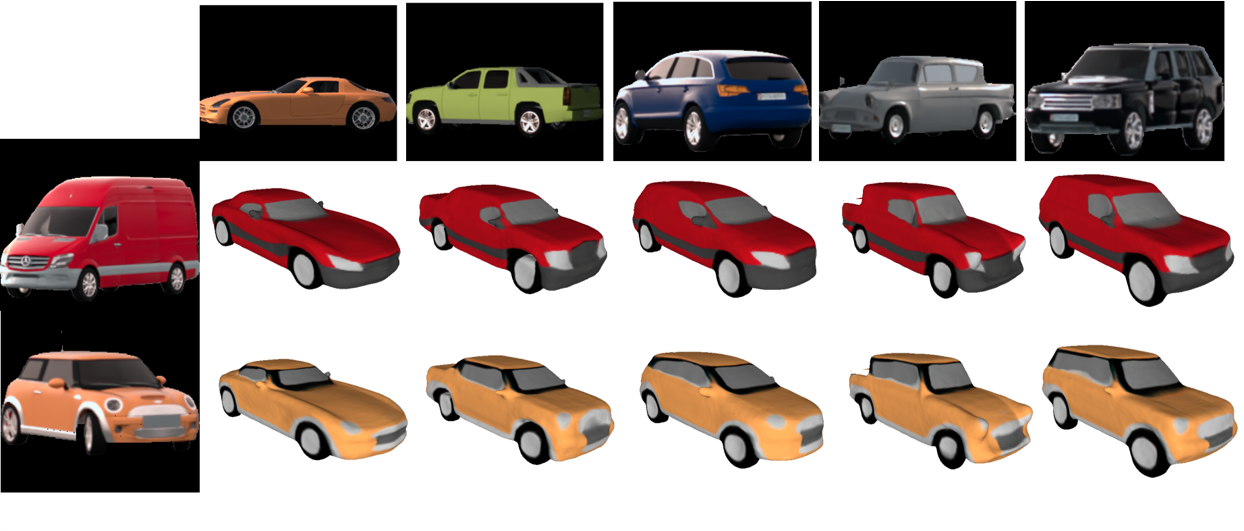}
\end{center}
\caption{Texture Transfer. By extracting shape latent codes from top row and texture latent codes from left col, our representation can freely couple shape and texture latent codes to generate new vehicle instances.}
\label{fig:tex_trans}
\end{figure}

\begin{figure*}[t]
\begin{center}
\includegraphics[scale=0.35]{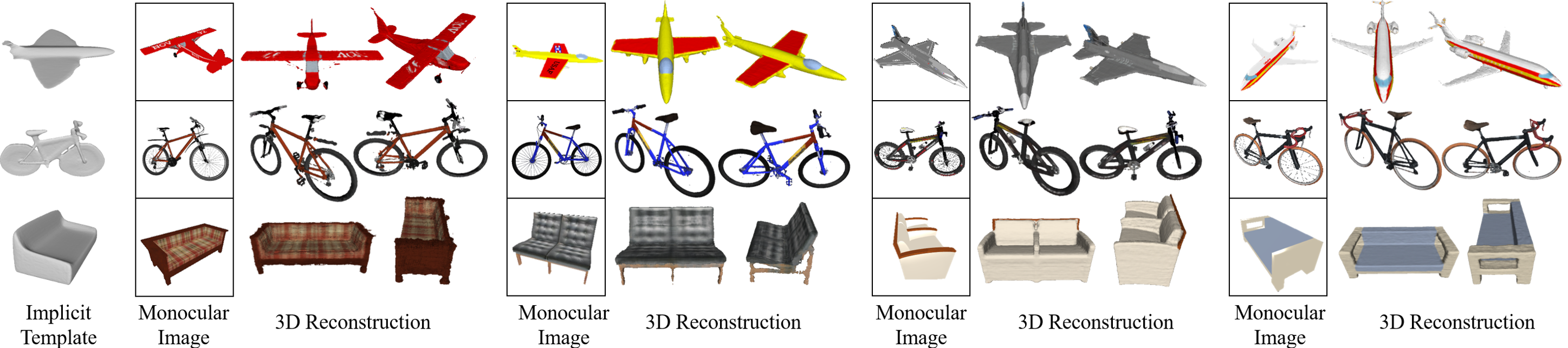}
\end{center}
\caption{Representation Generalization. Our model is extended to other object categories, with templates obtained from DIT.}
\label{fig:generalize}
\end{figure*}

\subsection{Evaluation}
\label{sec:evaluation}
\textbf{Evaluation on the disentanglement of representation.}
With the incorporation of the implicit semantic template mapping, our model avoid entangled representations and recover the surface texture in the 2-manifold of surfaces defined in the canonical template filed, thus supporting many downstream tasks such as texture transfer and editing. As shown in Fig.~\ref{fig:tex_trans}, we extract shape latent codes from top row to reconstruct the geometry and extract texture latent codes from the left col to recover the surface texture, resulting in plausible texture transfer, which proves the practicality of our disentanglement of geometry and texture latent spaces.

\begin{figure}
\begin{center}
\includegraphics[scale=0.25]{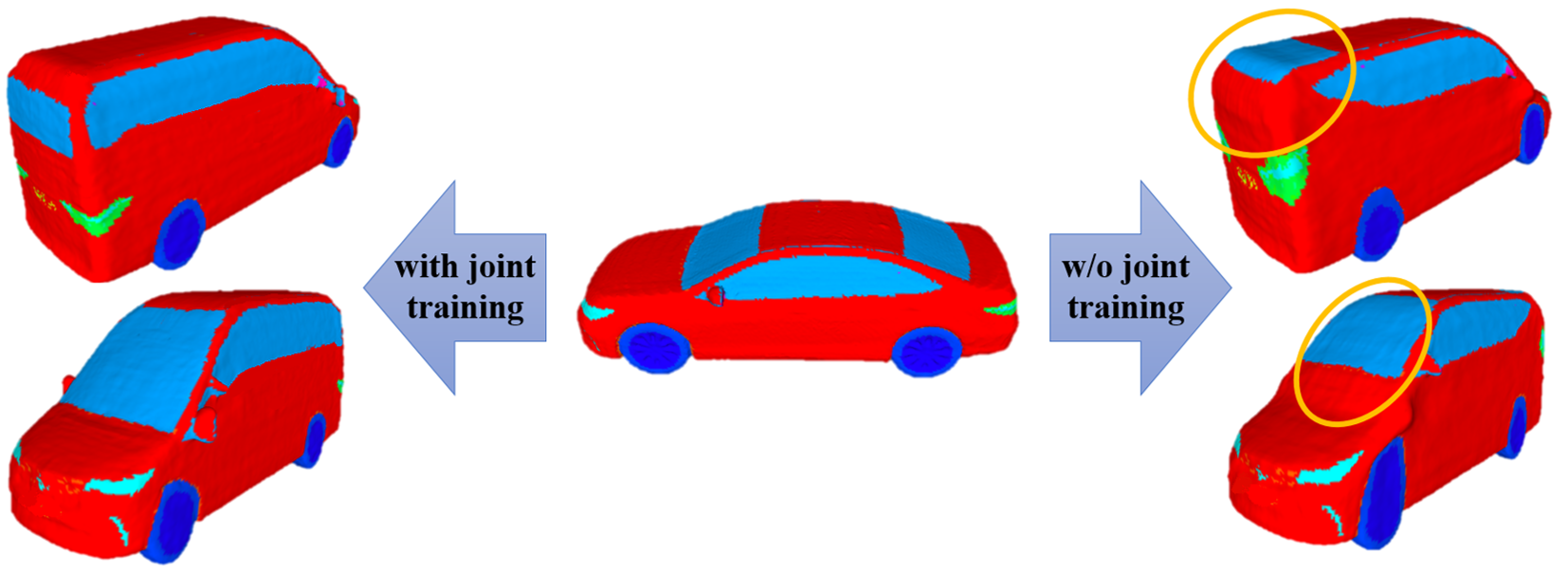}
\end{center}
\caption{Joint training strategy contributes to semantic template mapping. Note the semantic misalignment of front/back windows in the right column.}
\label{fig:cmp_trans}
\end{figure}

% \begin{table}
% \begin{center}
% \begin{tabular}{lcc}
% \hline
% & Without joint training & With joint training \\
% \hline
% RMSE &  5.259 & 0.7594 \\
% \hline
% \end{tabular}
% \end{center}
% \caption{Semantic correspondences accuracy evaluation on sparse keypoints annotation. Benefiting from the joint training strategy, we can establish meaningful semantic dense correspondences between various instances.}
% \label{table:kps}
% \end{table}

\textbf{Evaluation on the joint training strategy.}
% \subsection{Semantic Dense Correspondences}
Our method is able to establish semantic correspondences between different vehicle instances (right part of Fig.~\ref{fig:train_pipe}), attributed in our joint training strategy. We retrain a baseline network by firstly train the geometry branch and then train the texture branch conditioned on a fixed template mapping. As shown in Fig.~\ref{fig:cmp_trans}, without leveraging texture information as the guidance, the mapping process follows the principle of shortest distance to establish correspondences and ignores semantic information.
%The quantitative result in Tab.~\ref{table:idx} indicates that  the reduction in reconstruction quality. 
To evaluate the accuracy of semantic mapping, we utilize the key points annotation in our dataset and calculate the distance errors between the mapped key points and the target ones. RMSE score for comparison network is $\bm{5.259}$ while ours is $\bm{0.7594}$, which demonstrates that the joint training strategy helps establish meaningful semantic dense correspondences between various instances. 
Moreover, the decrease of numerical result in Tab.~\ref{table:idx} indicates that texture reconstruction quality does benefit from the semantic template mapping, implemented by geo-tex joint training strategy.

% \begin{figure}
% \begin{center}
% \includegraphics[scale=0.16]{multi_feature_fusion.png}
% \end{center}
% \caption{Validation of multi-scale features fusion for texture inference on real images. Note that by fusing both global and local information, our method can recover details such as car hubs and front lights and avoid noisy texture.}
% \label{fig:multi_feature_fusion}
% \end{figure}

% \textbf{Multi-scale Features Fusion for Texture Inference.} To demonstrate the validity of our multi-scale features fusion for texture inference, we train two baseline models using only global features and only local features respectively. As shown in Fig.~\ref{fig:multi_feature_fusion}, texture inference with only global features leads to consistent but coarse results, while the one only conditioned on local features could learn visible details but are more prone to noises. In contrast, our method of combining both global and local features solves these problems and produces consistent texture and captures fine-grain details as shown in the figure. 

% \begin{figure}
% \begin{center}
% \includegraphics[scale=0.29]{lc_inter2.png}
% \end{center}
% \caption{Latent Space Interpolations. Top row illustrates the continuous transformation of shape latent codes between two car instances. Bottom row illustrates the interpolation of texture latent codes extracted from two images. Note that results of texture space interpolation preserve semantic consistency.}
% \label{fig:lc_inter}
% \end{figure}

\textbf{Evaluation on illumination removal and material identification.}
%Complicated environment lighting conditions always pose challenges for intrinsic texture recovery. 
To alleviate the lighting effects of image appearance, we add an image-translation network to convert the input color images to albedo maps. The module effectively helps our network remove illumination and shading effects in 2D image domain and contributes to robust texture results. We retrain a comparison network by directly feeding original color images into texture encoders. As shown in Fig.~\ref{fig:de-illumination}, the network without the module tends to generate noisy results. 
Furthermore, though material identification based on monocular image is an ill-posed problem, with our implicit semantic template, the reconstructed intrinsic textured models obtain material parameters from the pre-designed template model and are able to generate realistic renderings through model relighting, as shown in the right part of the Fig.~\ref{fig:de-illumination}.

% \textbf{Generative Capability.} In this part, we show the generative capability of our latent space.  Fig.~\ref{fig:lc_inter} illustrates smooth interpolation in the shape and texture latent spaces. Specifically, in the case of texture latent space interpolation, for ease of visualization, we use only global latent codes and infer texture based on the geometry of template car. Moreover, as shown in Fig.~\ref{fig:tex_trans}, our representation allows for plausible texture transformation, which proves the practicality of our  disentanglement of shape and texture latent spaces.

% As shown in Fig.\ref{fig:lc_inter} and Fig.\ref{fig:tex_trans}, it's easy for our network to interpolate shape latent codes to generate new car instances and freely combine different shape and texture latent codes to transfer texture between different instances.
% Given two shape latent codes of different car instances, we interpolate them linearly and reconstruct geometry defined by the interpolated codes,  as shown in the bottom row in Fig.\ref{fig:generative_capability}. The texture of a given specific texture shows the interpolations in the shape latent space.

\textbf{Evaluation on the choice of the template.}
While our method select a sedan serving as the template car, to explore the sensitivity of our method to the choice of the template, we conduct three comparison experiments choosing an MPV, coupe and unit sphere as the template separately. For ease of comparison, we do not fuse local information for these experiments. Quantitative results are presented in Tab.~\ref{table:idx}. In general, our method is relatively insensitive to the template model and able to generate meaningful reconstruction results with different types of templates. We analyze that this arises from the fact that car shapes are almost homomorphic to the sphere, hence dense correspondences can be established for these template surfaces. Specifically, choosing the model close to the mean mesh within the category as the template will cause better performance. 
% Inspired by this, for experiments in \ref{subsec::Generalization}, we utilize the smooth template models learned from DIT~\cite{Zheng2021DIT} for each category.  

\begin{figure}
\begin{center}
\includegraphics[scale=0.3]{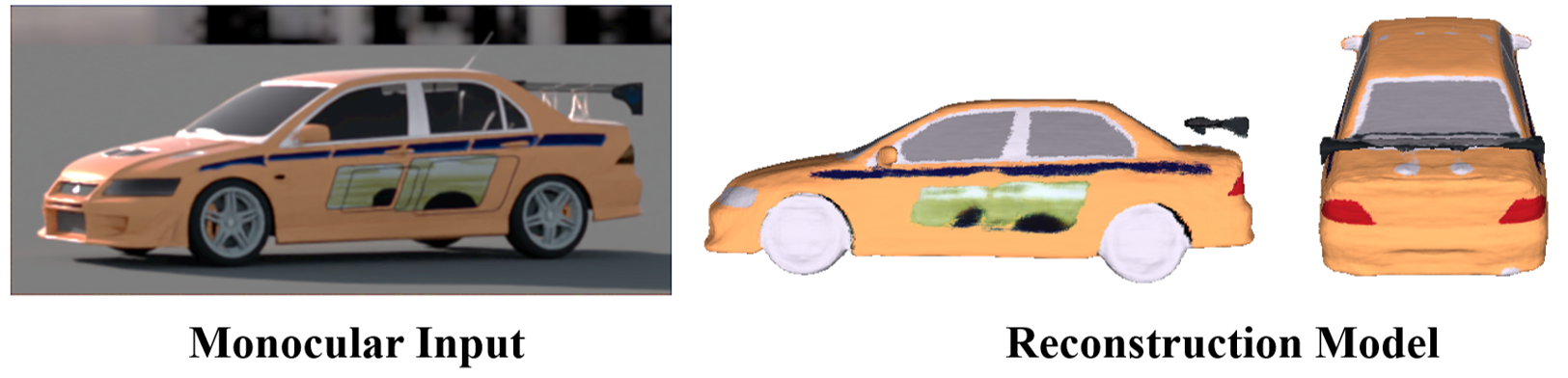}
\end{center}
\caption{Topology-varying vehicle reconstruction. Our method is able to represent vehicles with various typologies; see the rear spoiler of the reconstructed model. }
\label{fig:topology_vary}
\end{figure}

\textbf{Evaluation on topology-varying reconstruction.}
As the template mapping operator is defined in an implicit manner, our method preserves the advantage of implicit functions to represent topology-varying objects. Fig.~\ref{fig:topology_vary} presents an example of the reconstruction of a car with a separate rear spoiler. Results on other objects categories (Fig.~\ref{fig:generalize}) illustrates reconstruction cases with genus, the topology of which differs from the template meshes.

\subsection{Representation Generalization}
\label{subsec::Generalization}
In this section, we extend our representation to other object categories. We separately train our model on "sofa" and "airplane" category from ShapeNet~\cite{Chang2015ShapeNet}. To further prove our representation power, we also experiment on a collected bicycle dataset containing about 200 shapes. We use the template models learned by DIT~\cite{Zheng2021DIT} and conduct single-image 3D reconstruction experiments on these synthetic datasets. As shown in Fig.~\ref{fig:generalize}, our representation is qualified for generalization to other shape categories.

% \begin{figure*}
% \begin{center}
% \includegraphics[scale=0.5]{generative_capability.png}
% \end{center}
% \caption{Shape latent code interpolation and Texture Transfer}
% \label{fig:generative_capability}
% \end{figure*}

% \begin{figure*}
% \begin{center}
% \includegraphics[scale=0.4]{img/texture fusion strategy.png}
% \end{center}
% \caption{\color{red}{Texture fusion strategy. (a) Texture inference based on only local features is prone to noises. (b) By combing both global and local features, our method can reduce noises while preserving fine-grain details. However, resulting from the “imperfect” projection conditions, (c) misalignments of texture inference between visible and invisible parts is unavoidable. With our canonical implicit template filed, (d) a simple symmetric blend solution can solve this but cause blur result. With our proposed texture fusion strategy, our method can learn visible details as well as  produces global consistent texture.}}
% \label{fig:tex_fusion}
% \end{figure*}

%------------------------------------------------------------------------
\section{Conclusion}
\label{sec:conclusion}

In this paper, we have introduced VERTEX, a novel method for monocular vehicle reconstruction in real-world traffic scenarios. Experiments demonstrate that our method can recover 3D vehicle models with robust and detailed texture from a monocular image. Based on the proposed implicit semantic template mapping, we have presented a new geometry-texture joint representation to constrain texture distribution on the template surface, and have shown how to implement it with joint training strategy and a novel dataset.
%Additionally, by fusing the multi-scale features, our method can further generate stable 3D texture with fine-grained details. 
Moreover, we have demonstrated the advantages brought by the implicit semantic template to latent space disentanglement and material identification. We believe the proposed implicit geo-tex representation can further inspire 3D learning tasks on other classes of objects sharing a strong template prior. In future, we plan to extend our framework to handle the task of monocular video based vehicle reconstruction and leverage temporal information to improve the accuracy of texture estimation.
%, with deep implicit semantic template (DIST) representation and fusion of multi-scale features. 

{\small
\bibliographystyle{ieee_fullname}
\bibliography{egbib}
}

\end{document}

%%%%%%%%% TITLE
\title{VERTEX: {VE}hicle {R}econstruction and {TEX}ture Estimation \\From a Single Image Using Deep Implicit Semantic Template Mapping

– Supplementary Document –
}

\author{First Author\\
Institution1\\
Institution1 address\\
{\tt\small firstauthor@i1.org}
% For a paper whose authors are all at the same institution,
% omit the following lines up until the closing ``}''.
% Additional authors and addresses can be added with ``\and'',
% just like the second author.
% To save space, use either the email address or home page, not both
\and
Second Author\\
Institution2\\
First line of institution2 address\\
{\tt\small secondauthor@i2.org}
}

\maketitle
\appendix

% In this supplementary document, we provide more information about our implementation in Sec~\ref{sec:Implementation_Details} and additional experiments in Sec~\ref{sec::traffic_scene}. In the third part, we will Both our code and dataset will be made public upon the acceptance of the paper.

\section{Implementation Details}
\label{sec:Implementation_Details}
\subsection{Datasets}
\begin{figure}[h]
\begin{center}
\includegraphics[scale=0.8]{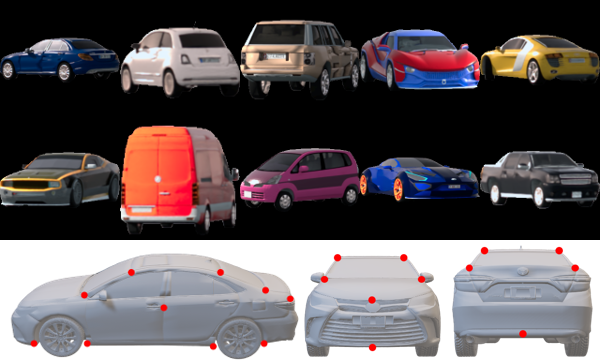}
\end{center}
\caption{Realistic synthetic dataset. Top: Samples of car instances rendered by the pbrt system. Bottom: Selected template car cad model labeled with 23 keypoints.}
\label{fig:dataset}
\end{figure}

Fig.~\ref{fig:dataset} illustrates some rendered images and the key-points annotation of our synthetic dataset.

\subsection{Network Architectures}
In this section, we introduce the architecture of each part of our network in detail.

As shown in the Fig.3 of the main paper, the backbone of VERTEX is composed of three main neural network blocks: \textbf{Template Mapping}, \textbf{RGB Decoder} and \textbf{Template SDF Decoder}.
%All of them are continuous functions represented by multi-layer perceptions (MLPs).

\textbf{Template Mapping} is based on a multi-layer perceptron, where the number of neurons is (512, 512, 512, 512, 256, 3) with weight normalization and ReLU activation except the last layer that uses the tanh function.
%Fig.~\ref{fig:template_mapping} illustrates the architecture of \textbf{Template Mapping}, composed of 4 fully-connected layers, each of which is applied with weight normalization and ReLU activation except the last layer that uses tanh function.
As the input and output of the module are both 3D points, we force the neural network to learn position offsets by adding a skip connection.
% \begin{figure}[h]
% \begin{center}
% \includegraphics[scale=0.21]{supp/mapping_decoder.png}
% \end{center}
% \caption{The architecture of \textbf{Template Mapping}. Conditioned on shape latent code $\bm{z}_{shape}$, the network translates the 3D point $p$ to corresponding position $p_{tp}$ in the template field.}
% \label{fig:template_mapping}
% \end{figure}

\begin{figure}[h]
\begin{center}
\includegraphics[scale=0.4]{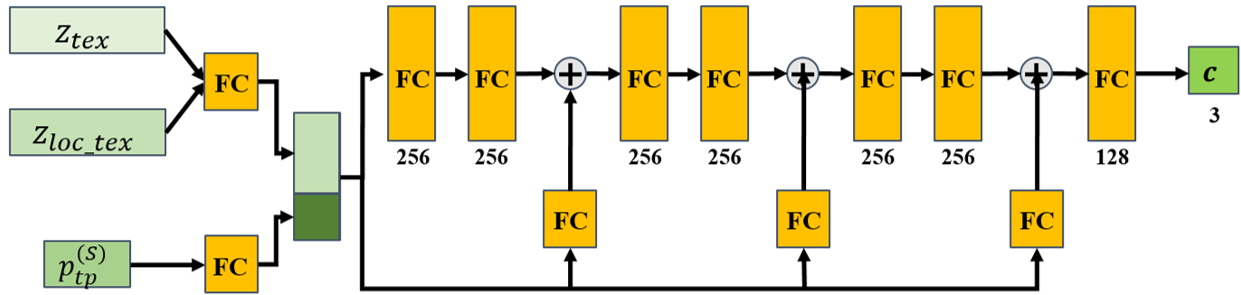}
\end{center}
\caption{The architecture of RGB Decoder. The network regresses color value $c$ for each template surface point $p_{tp}^S$.}
\label{fig::rgb_decoder}
\end{figure}

Fig.~\ref{fig::rgb_decoder} illustrates the architecture of \textbf{RGB Decoder}. We first fuse texture latent codes $\bm{z}_{tex}$ and $\bm{z}_{loc\_tex}(\bm{p})$ to form a conditional vector. Besides, instead of directly concatenating $3D$ position $\bm{p}_{tp}^{(S)}$ with $\bm{Z}$, we lift $\bm{p}_{tp}^{(S)}$ to an $128D$ vector to balance the dimensions of two input vectors. The backbone of \textbf{RGB Decoder} consists of 7 fully-connected layers with ReLU except the last layer that uses sigmoid activation. In spirit of \cite{2020PIFu}, we apply skip connections from the condition vector $\bm{Z}$ and input position $\bm{p}_{tp}^{(S)}$ to middle layers of MLP, by passing the concatenation through a fully-connected layer and adding it to the output of middle layers.

As the \textbf{Template SDF Decoder} only represents the template mesh, we choose a lightweight architecture of just 4 fully connected layers with a hidden size of 64. Through experimentation, we find that this configuration yields a good balance between reconstruction accuracy and net complexity. 
%To reduce the number of parameters to optimize, we pretrain this module and fix it during the end-to-end training.
 
% Fig.~\ref{fig:template_sdf_decoder} illustrates the architecture of \textbf{Template SDF Decoder}, and the setup of normalization and non-linear activations is similar to \textbf{Template Mapping}.

% \begin{figure}[h]
% \begin{center}
% \includegraphics[scale=0.25]{supp/template_sdf_decoder.png}
% \end{center}
% \caption{The architecture of \textbf{Template SDF Decode}r. The network embeds the signed distance field (SDF) of the template model, by querying the SDF value at the tempalte filed point $p_{tp}$.}
% \label{fig:template_sdf_decoder}
% \end{figure}

As for encoders, we use the ResNet-18 architecture~\cite{ResNet} to extract latent codes. We change the input dimension of original implementation in Pytorch~\cite{pytorch} to $4$ as we input four-channel RGB-Mask image. To extract local texture features $\bm{z}_{loc\_tex}$, we adopt the architecture of the image encoder for texture inference in ~\cite{2020PIFu}, which consists of 6 residual blocks and output 64-channel feature maps.   

\subsection{Training Procedure}
\paragraph{Albedo Recovery}
We utilize the U-net~\cite{ronneberger2015unet} to infer the intrinsic color in 2D domain. Noticing that training only with the L1 supervision term $L_{alb}$ may cause artifacts while handling real images, the training loss for the network additionally comprises four components:
\begin{equation}
    L = L_{alb} + L_{recon} + L_{percep} + L_{gan} + L_{fm}
    \label{eq13}
\end{equation}
where the $L_{recon}$ represents the L2 loss with the original image,  $L_{percep}$ represents the perceptual loss~\cite{2016Perceptual} with the ground truth albedo maps, $L_{gan}$ represents the GAN loss and $L_{fm}$ represents the feature matching loss~\cite{2017pix2pixHD} both of which constrain the network by means of the adversarial framework. Experimenting on real images, we find that such loss terms can reach the best performance and generate the most stable albedo maps.
%As the monocular albedo recovery is an ill-posed problem, to generate stable albedo maps,

\paragraph{End-to-end training details}
In our implementation, we set $\bm{z}_{tex}\in\mathbb{R}^{256}$ and $\bm{z}_{loc\_tex}\in\mathbb{R}^{64}$. For the class of vehicles, $\bm{z}_{shape}\in\mathbb{R}^{32}$; for other object categories, $\bm{z}_{shape}\in\mathbb{R}^{128}$. Specifically, we normalize $\bm{z}_{shape}$ by projecting the latent vector onto the unit hypersphere. The weights of Eq. 7 in the main paper are: $w_{g}=10.0$, $w_{reg}=0.1$, $w_{t}=100.0$. We use the Adam optimizer \cite{2014Adam} with a fixed learning rate of $1\times{10^{-4}}$, the batch size of 32, the number of epochs of 500, the number of volume sampled points of 8000, and the number of surface sampled points of 6000 per object in every training batch. The training takes 2 days on two 1080ti GPUs.

\subsection{Inference}

During inference, to obtain the proper local texture latent codes $\bm{z}_{loc\_tex}$ for surface points, we assume that the observed image is the central projection of normalized model with regard to a perspective camera with a fixed FOV of $Fov=10^{\circ}$.
The initial distance is set as 10$m$. We use the publicly available differentiable renderer from ~\cite{kato2018neural}. With the model scale fixed, we optimize 6D pose by minimizing 2D silhouette loss and establish projective correspondences between 3D surface points and feature maps.

To further improve the texture quality in unobserved region, we can adopt a texture fusion strategy to refine the global texture latent $\bm{z}_{tex}$ and the parameters of the RGB Decoder. Concretely, we minimize the difference between the regressed texture conditioned on $\bm{z}_{tex}$ and the fine texture conditioned on both $\bm{z}_{loc\_tex}$ and $\bm{z}_{tex}$, as well as the symmetry loss and the regularization loss to ensure global texture consistency.

% \caption{\color{red}{Texture fusion strategy. (a) Texture inference based on only local features is prone to noises. (b) By combing both global and local features, our method can reduce noises while preserving fine-grain details. However, resulting from the “imperfect” projection conditions, (c) misalignments of texture inference between visible and invisible parts is unavoidable. With our canonical implicit template filed, (d) a simple symmetric blend solution can solve this but cause blur result. With our proposed texture fusion strategy, our method can learn visible details as well as  produces global consistent texture.}}

% \section{Additional Results}
% \label{sec:Additional_Results}
% \subsection{Additional Experiments}
% In the main paper, we show comparisons with PIFu~\cite{2020PIFu} and Onet + Texture Field \cite{Mescheder2019Occupancy,Oechsle2019Texture}. For fair comparison, different from original implementations, we adjust input to 4-channel RGB-M image , and for Onet + TF, we set the geometry latent code as $32D$ latent vector and texture latent code as $256D$ latent vector, the dimensions of which are equal to those in our method. In this section, we retrain the contrast works with higher-dimensional latent codes (for Onet + TF, $\bm{z}_{shape}\in\mathbb{R}^{512}$ and $\bm{z}_{texture}\in\mathbb{R}^{512}$; for PIFu, $\bm{z}_{local\_shape}\in\mathbb{R}^{256}$ and $\bm{z}_{local\_texture}\in\mathbb{R}^{256}$) and show comparison results in Fig.~\ref{fig::compare}.

\section{Further Experiments}
\label{sec:sup_experiments}
\subsection{Multi-scale Features Fusion for Texture Inference}
\begin{figure}[h]
\begin{center}
\includegraphics[scale=0.16]{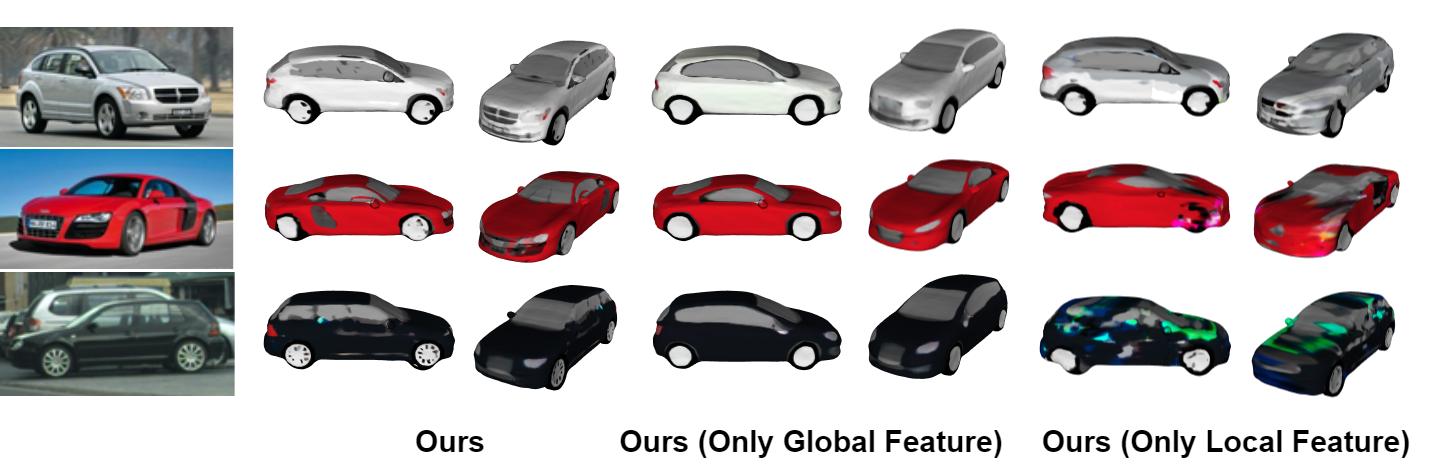}
\end{center}
\caption{Validation of multi-scale features fusion for texture inference on real images. Note that by fusing both global and local information, our method can recover details such as car hubs and front lights and avoid noisy texture.}
\label{fig:multi_feature_fusion}
\end{figure}

To demonstrate the validity of our multi-scale features fusion for texture inference, we train two baseline models using only global features and only local features respectively. As shown in Fig.~\ref{fig:multi_feature_fusion}, texture inference with only global features leads to consistent but coarse results, while the one only conditioned on local features could learn visible details but are more prone to noises. In contrast, our method of combining both global and local features solves these problems and produces consistent texture and captures fine-grain details as shown in the figure. 

\subsection{Latent Space Interpolations}
\begin{figure}[h]
\begin{center}
\includegraphics[scale=0.29]{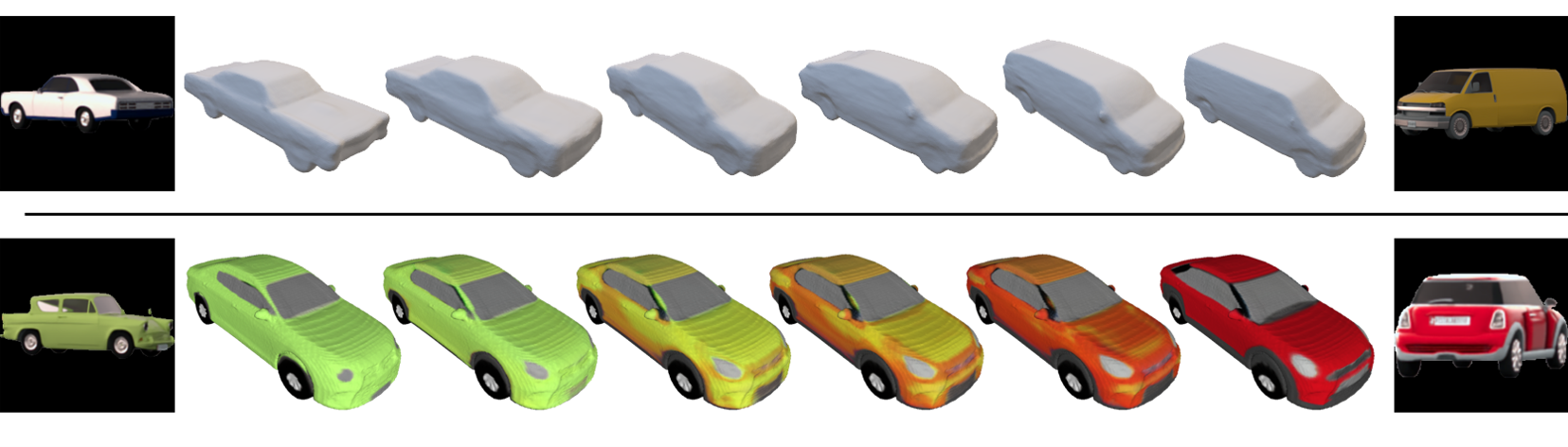}
\end{center}
\caption{Latent Space Interpolations. Top row illustrates the continuous transformation of shape latent codes between two car instances. Bottom row illustrates the interpolation of texture latent codes extracted from two images. Note that results of texture space interpolation preserve semantic consistency.}
\label{fig:lc_inter}
\end{figure}

In this part, we show the generative capability of our latent space.  Fig.~\ref{fig:lc_inter} illustrates smooth interpolation in the shape and texture latent spaces. Specifically, in the case of texture latent space interpolation, for ease of visualization, we use only global latent codes and infer texture based on the geometry of the template car.

% \section{Field Visualzation}
% In this section, we aim at explaining the differences between our representation and Texture Field (TF)~\cite{Oechsle2019Texture}. Similar to TF, we try to visualize what the  
% \begin{figure}[h]
% \begin{center}
% \includegraphics[scale=0.45]{tf_cmp.png}
% \end{center}
% \caption{In this figure, we use the example of texture transfer to explain the difference between our representation and Texture Field. To clearly demonstrate that we successfully represent the texture distribution onto the template surface, we retrain the network with an additional loss: constrain 3D points far away from the surface to regress fixed RGB value (grey). As figure shows, with such additional constraint our network still produces high-quality texture while TF diffusing the surface texture into 3D space might cause blur results. }
% \label{fig:tf_cmp}
% \end{figure}

% \begin{figure*}
% \begin{center}
% \includegraphics[scale=0.3]{supp/Compare2.png}
% \end{center}
% \caption{Comparison results on in-the-wild images. Monocular input images are shown in the left column. Two render views different from the original observation are provided to demonstrate reconstruction quality. Our results still achieve better performance in terms of both robustness and accuracy, compared with contrast works with higher-dimensional latent variables.}
% \label{fig::compare}
% \end{figure*}

\subsection{Results on traffic scenarios}
\begin{figure*}
\begin{center}
\includegraphics[scale=0.5]{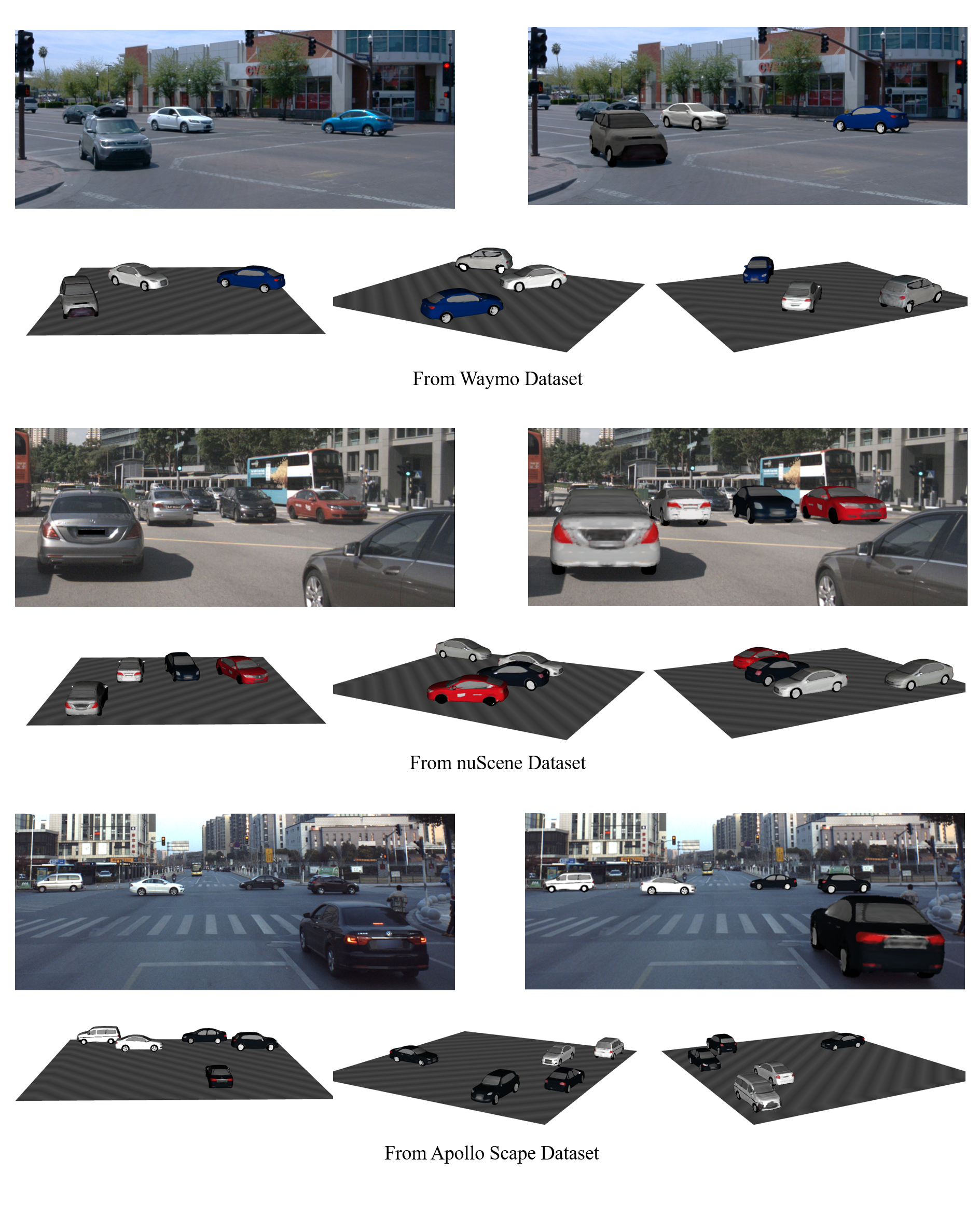}
\end{center}
\caption{Results on real-world traffic scenarios. For each case, with the monocular input RGB image (shown in the left top), we visualize our reconstruction results by reprojecting them to the image plane (shown in the right top) and presenting three novel views of the reconstructed scene (shown in the bottom row).}
\label{fig::scene}
\end{figure*}

We present more reconstruction results on real-world traffic scenarios from diverse datasets~\cite{wang2019apolloscape, waymo, nuscenes2019} in Fig.~\ref{fig::scene}.

\section{Limitation}
% Our method is able to recover realistic 3D vehicle model from a single-view natural image.
Extreme lighting and shading effects and crowd mutual occlusion in real images will pose challenges for our current work. Our method is competent to recover realistic 3D vehicle model under general outdoor environments. However, overexposure and intricate shadows in images of outdoor scene makes it difficult to recover intrinsic color for monocular reconstruction methods, as it's hard to distinguish illumination effects and local texture detail. Besides, in real-world settings, occlusions are common and only a part of the vehicle is captured in the observed image, which has not yet been taken into consideration for our current work. To handle these challenges in future, we plan to leverage multi-frame observations and temporal information by extending our framework to the monocular video based vehicle reconstruction.

{\small
\bibliographystyle{ieee_fullname}
\bibliography{egbib}
}